\documentclass[letterpaper]{article} 
\usepackage{aaai24}  
\usepackage{times}  
\usepackage{helvet}  
\usepackage{courier}  
\usepackage[hyphens]{url}  
\usepackage{graphicx} 
\usepackage{tabularx}
\usepackage{natbib}  
\usepackage{caption} 
\usepackage{algorithm}
\usepackage{algorithmic}
\usepackage{amsmath}
\usepackage{graphicx}
\usepackage{newfloat}
\usepackage{listings}
\usepackage{multirow}
\usepackage{newfloat}
\DeclareCaptionStyle{ruled}{labelfont=normalfont,labelsep=colon,strut=off} 
\floatstyle{ruled}
\newfloat{listing}{tb}{lst}{}
\floatname{listing}{Listing}
\title{LLM Based Multi-Document Summarization \\Exploiting Main-Event Biased Monotone Submodular Content Extraction}
 \author{Litton J Kurisinkel,  Nancy F. Chen \\
         Institute for Infocomm Research, A*STAR, Singapore\\
         litton\_kurisinkel, nfychen@i2r.a-star.edu.sg\\
         }

\usepackage{bibentry}

\begin{document}

\maketitle

\begin{abstract}
Multi-document summarization is a challenging task due to its inherent subjective bias, highlighted by the low inter-annotator ROUGE-1 score of 0.4 among DUC-2004 reference summaries. In this work, we aim to enhance the objectivity of news summarization by focusing on the main event of a group of related news documents and presenting it coherently with sufficient context. Our primary objective is to succinctly report the main event, ensuring that the summary remains objective and informative. To achieve this, we employ an extract-rewrite approach that incorporates a main-event biased monotone-submodular function for content selection. This enables us to extract the most crucial information related to the main event from the document cluster. To ensure coherence, we utilize a fine-tuned Language Model (LLM) for rewriting the extracted content into a coherent text. The evaluation using objective metrics and human evaluators confirms the effectiveness of our approach, as it surpasses potential baselines, demonstrating excellence in both content coverage, coherence, and informativeness.
\end{abstract}

\section{Introduction}
Presenting information in text format has been critical to the development of human civilizations. Thus, text generation is an important field in artificial intelligence and natural language processing, where the input to such natural language generation models could take the form of text, graphs, images, or database records \cite{zhang2020pegasus,song2018graph,wiseman2018learning,zhang2020radiology}. 

The pursuit of text-to-text generation in the NLP community has taken various approaches and seen significant advancements, especially with the advent of large language models \cite{grail2021globalizing}. In the past, certain methods have treated text-to-text generation as a search problem and utilized discrete optimization techniques and greedy algorithms to find solutions \cite{barzilay2006aggregation,clarke2008global}. To improve the consistency of greedy methods, some researchers have incorporated monotone submodular functions \cite{lin2011class}.

Operating in the discrete space has given these methods an advantage in terms of exerting more control during text generation \cite{erten2021ontology}. However, they suffer from a critical drawback – the inability to effectively learn from a large volume of training data and produce text that truly resembles human-written text with good style.

 Data-driven text generation techniques based on neural networks heavily rely on training datasets to build and fine-tune their models \cite{xiao2021primera}. The reliability and efficacy of these models are inherently tied to the quality and credibility of the datasets employed \cite{jain2020overview}. In the domain of text summarization, many researchers have utilized autocreated datasets \cite{fabbri2019multi}. Nevertheless, it is essential to acknowledge the mounting concerns and potential criticisms regarding the reliability of presently available datasets, particularly in the context of multi-document summarization \cite{wolhandler2022multi}. 

Large language models are inherently designed to learn from large amounts of training data using computations in a continuous space with dense representations to produce text that cannot be discriminated from human-written text \cite{touvron2023llama}. However, memory-efficient LLMs suffer from constraints on context length, which restricts their direct use for multi-document summarization \cite{li2023unlocking}. Also, unlike traditional methods, neural networks inherently provide fewer provisions to control intermediate computations \cite{erdem2022neural}. The factor of controllability is crucial when it comes to real-world applications of text generation methods. Additionally, lightweight large language models are yet to overcome the curse of hallucinations \cite{azaria2023internal}. In such a situation, extract-rewrite methods can be a relevant option to explore.

The inter-annotator ROUGE score of 0.4 between DUC-2004 reference summaries may indicate the need for more objective methods in defining multi-document summarization techniques \footnote{https://duc.nist.gov/duc2004/}. Our system aims to summarize a cluster of documents related to the same main event. The objective of our summarization is to report the main event with the most relevant information, providing the necessary context to infer the main event coherently. Our objective is conditioned by previous observations on event-based discourse analysis for news articles \cite{can2013news}. 

In a nutshell, the contributions of the paper are:
\begin{itemize}
    \item A main event biased greedy method which formulates a more objective content extraction method for Multi-Document Summarization (MDS) using a monotone submodular function with linear components for coverage, diversity, and coherence.
    \item A fine-tuned Language Model (LLM) which takes the extracted content and re-writes it to create a coherent summary.
\end{itemize}
The paper also introduces an annotated test set consisting of 30 clusters of documents and corresponding main event-focused summaries.
\begin{figure*}[!t]
  \centering
\includegraphics[width=0.8\textwidth]{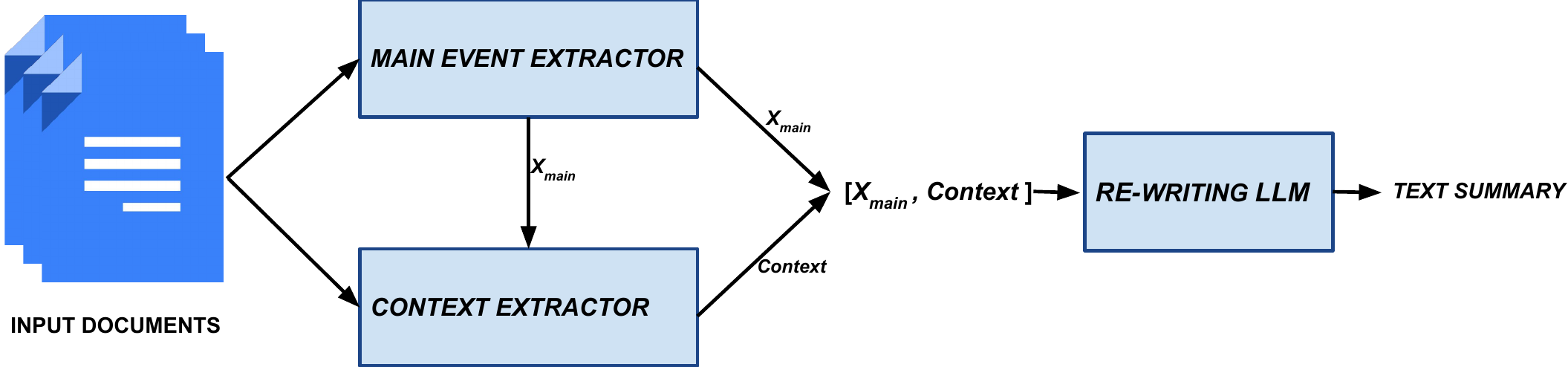} 
  \caption{Main Event biased Multi- Document Summarization: Context Extractor is a Monotone Sub Modular Function}
  \label{f:appraoch3}
\end{figure*}
\section{Related Work}
In the field of text summarization, there are two main approaches: extractive and abstractive methods. Extractive methods involve directly selecting and combining sentences from the original text, often yielding fluent outputs as they use human-written content \cite{ lin2011class, wang2008multi}. Some extractive approaches have used parsed sentence structures to remove noise and extract relevant content for the summary \cite{morita2013subtree}. To move towards abstractive summarization, researchers introduced sentence compression techniques, which aimed to reduce noise and condense information \cite{lin2003improving, zajic2006sentence,almeida2013fast}. However, these methods faced challenges in maintaining coherent reading for the summary reader. To address the coherence issue, \citet{christensen2013towards} proposed a method that leverages corpus-level discourse graphs to achieve both structural and topical coherence. During summary extraction, their system attempts to balance salience (important content) and coherence. Despite their efforts in enhancing coherence, these approaches  have limitations when it comes to effectively leveraging large datasets and the learning capabilities of neural networks.

Numerous recent studies have developed text-to-text generation methods based on neural networks \cite{Rush2015neural, Chopra2016abstractive, Zhong2020extractive, Liu2019fine}. Some of these works focus on generating summaries from input documents \cite{see2017get, wang2019learning, liu2019text, zhang2020pegasus}. The fundamental concept involves training a neural network to automatically extract syntactic and semantic features from the input text to generate the desired output. However, these methods are less tractable and provide limited control over summary attributes. Addressing these challenges will be crucial for advancing the efficacy and versatility of text-to-text generation techniques.

Recent interest in NLP methods is sparked by the introduction of Large Language models (LLMs) \cite{touvron2023llama,penedo2023refinedweb}. These language models possess a large set of parameters, equipping them with exceptional learning capabilities. There are LLMs specifically trained for various NLP tasks, including text-to-text generation and machine translation \cite{ni2021sentence}. However, memory-efficient LLMs face limitations due to their low context length \cite{chen2023extending} and may still experience hallucinations \cite{azaria2023internal}. In this current work, we explore a more manageable approach that can be applied in real-world scenarios. This approach involves using a dedicated content extraction scheme to produce coherent output summaries in a guided manner by rephrasing the extracted content.

\section{Problem Defenition}
We define our problem in 3 steps.
\begin{align*}
    x_{\text{main}} &= E_{\text{main}}(X) \\
    \text{context}({x_{\text{main}}}) &= F(X, x_{\text{main}}) \\
    S_{\text{sum}} &= Z([x_{\text{main}}, \text{context}({x_{\text{main}}})])
\end{align*}
Where $X$ represents the input set of documents that are related to a main event which is created through event-based clustering. The function $E_{\text{main}}$ is specifically designed to extract a sentence $x_{\text{main}}$ from the collection $X$, and this sentence is intended to represent the main event. Afterwards, the function $F$ takes on the task of extracting a set of sentences referred to as $\text{context}(x_{\text{main}})$. These extracted sentences provide the most relevant contextual information for $x_{\text{main}}$, helping to give a better understanding of the main event. To generate the final summary $S_{\text{sum}}$, we employ the re-writing model $Z$. This model takes the concatenation of the main event sentence $x_{\text{main}}$ with its corresponding context $\text{context}(x_{\text{main}})$ to rewrite the information contained in them to a final coherent summary. 
\section{Method}
Our method involves three main steps: main event extraction, context extraction, and re-writing. These steps are depicted in Figure \ref{f:appraoch3}. We rely on an event discourse-based approach to implement the main event extractor $E_{\text{main}}$. The main event context extractor $F$ is a greedy method based on an unsupervised monotone submodular function, while the re-writing model $Z$ is a comparatively lightweight, large language model fine-tuned for this purpose. We will explain each of these steps in detail in the subsequent subsections.
\subsection{$E_{\text{main}}$: 
Main Event Extraction}
To extract the main event from articles, we rely on the discourse structure proposed by \citet{can2013news}, which presents a main-event centric view of discourse. According to this approach, all linguistic units in a document share a discourse relation with the linguistic unit that represents the main event. \citet{choubey2020discourse} studied the case where linguistic units are sentences and introduced a method to identify discourse labels. We adopted the same method to identify and extract main event sentences from a document.
To determine the main event of input set of documents X, we create a candidate set of sentences by extracting main event sentence from each document in the input set. From these candidates, we select the one with the maximum coverage value as the main event of the document X as follows.
\begin{align}
x_{\text{main}} &= \operatorname*{argmax}_{x_{i} \in \text{candidates}
} C(\{x_{i}\})
\end{align}
$C(x_{i})$ is the coverage of the sentence $x_{i}$ which is computed using Equation \ref{coverage_eqn}, and details of it will be explained in subsequent sections.

\subsection{F: A Greedy Method for Main event Context Extraction}
The context extractor, denoted as F, focuses on extracting the most relevant information needed to provide necessary context for the main event $x_{\text{main}}$. F is an extractive summarization method that is biased towards the main event. The problem of extractive summarization can be represented as an argument maximization operation as follows.
\begin{align}
S_{\text{opt}} &= \underset{S}{\operatorname{argmax}}\, F(S)
\end{align}
Where $S_{\text{opt}}$ is the optimum extractive summary and $F$ is a function defined in the space of candidate summaries. \citet{lin2011class} comes out with the interesting observation that if $F$ is a monotone submodular function, there exists a greedy method that can approximate the optimum summary by a factor of 0.692.
\begin{equation}
\begin{split}
F(S_{\text{GREEDY}}) &\geq \left(1 - \frac{1}{e}\right)F(S_{\text{opt}}) \\
&\approx 0.632F(S_{\text{opt}})
\end{split}
\end{equation}
Where $S_{\text{GREEDY}}$ is a summary computed using corresponding greedy method. Inspired by this relation, we formulate a $F$ as follows. 
\begin{align}
F(S) &= C(S) + \lambda_1 \cdot D(S) + \lambda_2 \cdot B_{\text{main}}(S)
\end{align}\\
Where $C$, $D$, and $B_{\text{main}}$ are functions that measure the topical coverage, topical diversity, and main event-based importance, respectively, of candidate summary $S$. Ensuring that these linear components are monotone submodular will consequently ensure that $F$ is submodular. The rest of this section explains each one of them in detail. 
\subsubsection{$C(S)$ : Content Coverage} 
$C(S)$ is the set function that estimates the extent to which a candidate summary $S$ is a topical representative of the input set of documents $X$. We use an enhanced version of the topical coverage function proposed by \citet{lin2011class}, which is based on dense sentence representations, to estimate coverage.
\begin{align}
C(S) &= \sum_{i \in U} \min \{c_i(S), \alpha * c_i(U)\}
\label{coverage_eqn}
\end{align}
where $\alpha$ is an empirically optimized constant $U$ represents the set of all sentences in the input set of documents and,
\begin{align}
c_i(S) = \sum_{j \in S} Sim(i,j)
\end{align}
Where $Sim$ computes the cosine similarity between S-BERT representations of argument sentences \cite{reimers2019sentence}. 
\subsubsection{$D(S)$:Content Diversity}
Our diversity function, designed to measure the content diversity of candidate summaries, is a variant of the diversity function proposed by \citet{lin2011class}. The process involves clustering the input set of sentences using the affinity propagation algorithm \cite{dueck2009affinity} with the similarity metric $Sim$. We then utilize the information regarding these clusters to compute the content diversity of summaries as follows.
\begin{align}
D(S) = \sum_{i=1}^{K}\sqrt{ \sum_{j \in P_i \cap S} \frac{c_i(U)}{N}
}
\end{align}
Where $N$ be the total number of sentences in $U$, and let $P_i$ be the set of sentences in cluster $i$. Thus, the diversity function encourages summaries to have representations for more number of clusters.
\begin{table*}[t]
  \centering
  \small
  \begin{tabular}{lcccc}
    \hline
    \textbf{Dataset} & \textbf{Model} & \textbf{ROUGE-1} & \textbf{ROUGE-2} & \textbf{ROUGE-L} \\
    \hline
    \multirow{6}{*}{Multi- News} & PEGASUS &  36.5  &  10.5 & 18.7 \\
    & BART& 27.3 & 6.2 & 15.1 \\
    & LED& 17.3 & 3.7 & 10.4\\
     & PRIMERA& 42.0 & 13.6 & 20.8 \\
     & $C(S) + \lambda_1 \cdot D(S)$ & 40.5 & 12.7 & 19.3 \\
     & $C(S) + \lambda_1 \cdot D(S) + \lambda_2 \cdot B_{\text{main}}(S)$& \textbf{43.0} & \textbf{14.2} & \textbf{21.3} \\
    \hline
    \multirow{6}{*}{WCEP} & PEGASUS & 33.2 & 12.7 & 23.8 \\
    & BART& 20.2 & 5.7 & 15.3 \\
    & LED& 18.8 & 5.4 & 14.7 \\
     & PRIMERA&  28.0 &  10.3 &  20.9 \\
     & $C(S) + \lambda_1 \cdot D(S)$& 32.5 & 11.3 & 20.7 \\
     & $C(S) + \lambda_1 \cdot D(S) + \lambda_2 \cdot B_{\text{main}}(S)$ & \textbf{33.7} & \textbf{12.9} & \textbf{23.7} \\
    \hline
    \multirow{6}{*}{Main Event}& PEGASUS &  32.1  &  9.5 & 16.1 \\
    & BART& 25.3 & 6.1 & 14.0 \\
    & LED& 16.0 & 4.1 & 9.2\\
     & PRIMERA& 42.3 & 12.8 & 19.1 \\
     & $C(S) + \lambda_1 \cdot D(S)$ & 41.3 & 13.6 & 20.5 \\
     & $C(S) + \lambda_1 \cdot D(S) + \lambda_2 \cdot B_{\text{main}}(S)$ & \textbf{49.9} & \textbf{22.7} & \textbf{27.9} \\
    \hline
  \end{tabular}
  \caption{Text Summarization Performance Comparison using ROUGE Metrics: Our Method with Main-Event Bias Outperforms Peer Systems}
  \label{tab:rouge_results}
\end{table*}
\subsubsection{$B_{\text{main}}(S)$:Main Event Based Relevance}
\label{main_event_bias}
Our approach mainly focuses on reporting the main event with essential details in the summary. Therefore, $B_{\text{main}}$ is a crucial component that computes the relevance of a candidate summary with respect to the main event. One way to compute this could be based on the discourse label of sentences in the input document with respect to the main event, as per the news discourse structure suggested by \cite{choubey2020discourse}. This approach ensures that the summary emphasizes the main event and includes the necessary information related to it.
\begin{align}
  B_{\text{main}} &= \sum_{s \in S} \text{P}\left(s/\text{discourse}(s,x_{\text{main}})\right)
\end{align}
Where $\text{P}$ computes the probability of sentence $s$ being in the summary with respect to its discourse label $\text{discourse}(s, x_{\text{main}})$. However, it is extremely difficult to identify these discourse labels in the context of multi-document, as it requires extensive effort. For this reason, we investigated an approach that could implicitly compute the relevance of a sentence with respect to the main event without explicitly identifying discourse labels. Consequently, we found a method based on co-occurrence probability, in which main event-based relevance is computed as follows.
\begin{align}
    B_{\text{main}}(S) = \sum_{i \in S} Coc(i,x_{\text{main}})
\end{align}
Where $Coc$ is a neural network that measures the co-occurrence probability of argument sentences. Given sentence vectors $s_{1}$ and $s_{2}$ for argument sentences, the network computes $[s_{1}, s_{2}]$, $s_{1}-s_{2}$, $s_{1} * s_{2}$, and $|s_{1} - s_{2}|$ \cite{xu2019cross}. These features are concatenated and fed into an MLP layer. The model is made order-agnostic by training a forward model with input $(s_{1}, s_{2})$ and a backward model with input $(s_{2}, s_{1})$, using the same architecture but different parameter sets. The co-occurrence score is computed as the average of the two models. The loss for a triplet $\mathcal{T} = (s_{i}, s_{p}, s_{n})$ with positive pair $(s_{1}, s_{p})$ and negative $(s_{1}, s_{n})$ is computed as follows,  
\begin{align}
\mathcal{L}_{\mathcal{T}} = \max(0, m + Coc(\mathbf{s_{1}}, \mathbf{s_{p}}) - Coc(\mathbf{s_{1}}, \mathbf{s_{n}}))
\label{co_occurance_loss}
\end{align}
Where $m$ is the margin.
\subsection{Summarization}
\citet{lin2011class} deductively showed that $C(S)$ and $D(S)$ are monotonically submodular. $B_{\text{main}}(S)$ is monotonically submodular as $Coc(i, x_{\text{main}})$ is constant for all sentences in the input document set. Consequently, $F$ is monotonically submodular. We start our corresponding greedy method with a maximum weight selection scheme to extract $S_{\text{GREEDY}}$, beginning with the singleton set ${x_{\text{main}}}$. This approach allows the extracted summary $S$ to effectively approximate [$x_{\text{main}}, \text{context}({x_{\text{main}}})$]. The number of sentences to be extracted, $N$, is decided using the variance $\sigma^2$ among similarity values between input sentences measured using $sim$.

\begin{align*}
N=&\lfloor k + c \cdot \sigma^2 \rfloor
\end{align*}

\noindent Where $k$ and $c$ are constants which are optimized emprically.
\subsection{Z: Re- writing Model}
The extracted set of summary sentences needs to be rewritten into a coherent text using a large language model fine-tuned for this purpose. We created a parallel dataset consisting of the extracted set of summary sentences and their corresponding coherently written versions using the method from the multi-news dataset \cite{fabbri2019multi}. To create the dataset, we used the alignment method employed by \citet{wolhandler2022multi}. We fine-tuned the flan-t5-xl model\footnote{https://huggingface.co/google/flan-t5-xl}, which is specifically pre-trained for text-to-text generation, and introduced the 're-write' prompt during training. The T5 model is lightweight and requires less compute for training and inference.
\section{Experiments}
 We conducted experiments to evaluate various attributes of extracted summaries, such as content coverage, and coherence. Additionally, we evaluated our submodels, which estimate the sentence co-occurrence score of sentences and rewrite the extracted summary. This section provides further details about our experimental settings and the datasets used for evaluations.
\subsection{Co-Occurance Model}
We collected 45,000 news documents and extracted the first three sentences from each document to constitute our pre-training dataset. Subsequently, we trained the model using multi-news summaries \cite{fabbri2019multi}. Any pair of documents taken from the same piece of text is a positive pair, and any pair of sentences taken from two different texts represents a negative pair. For each positive pair, we choose a corresponding negative pair to form the triplet explained in Equation \ref{co_occurance_loss}. The cardinality of our training, development, and test sets is 45,000, 1800, and 1800, respectively. We set the margin $m$ to 5.4 in Equation \ref{co_occurance_loss}. The results are shown in Table \ref{co_occurance_results}. 0.77 serves as a dependable F-Measure value suitable for use as a sub-component.
\begin{table}
    \centering
    \small
    \label{tab:metrics}
    \begin{tabular}{|c|c|c|}
        \hline
        \textbf{Precision} & \textbf{Recall} & \textbf{F-Measure} \\
        \hline
        0.93& 0.70 & 0.77 \\
        \hline
    \end{tabular}
    \caption{Sentence Co- Occurance Results}
    \label{co_occurance_results}
\end{table}
\subsection{Fine-tuning LLM for re- writing} 
We use the flan-t5-xl model\footnote{https://huggingface.co/google/flan-t5-xl} for the purpose of re-writing. As explained earlier in the paper, we created a dataset by aligning summary sentences in the Multi-news dataset to source sentences. We created around 3000 parallel records for fine-tuning, 450 for development, and 450 for testing. The fine-tuned re-writing model yielded a BLEU score of 0.30.
\subsection{Text Summarization}
We conducted experiments to evaluate our summaries on the dimensions of content coverage and coherence. We relied on objective metrics such as ROUGE-1, ROUGE-2 and ROUGE-L for estimating content coverage while on human evaluation for estimating coherence.
\subsubsection{Data:} We evaluated our approach for summarization using public datasets such as Multi-news \cite{fabbri2019multi} and WCEP test sets \cite{ghalandari2020large}. Multi-news contains a cluster of related news documents as input and corresponding summaries. The WCEP dataset for multi-document summarization (MDS) consists of short, human-written summaries about news events, obtained from the Wikipedia Current Events Portal (WCEP), each paired with a cluster of news articles associated with an event. In addition to these datasets, we created a dataset in which summaries are written to report the main event. The particular dataset consists of 30 clusters of documents. Each cluster of documents is related to a single main event and contains two to four documents. Our annotators are two post-graduate research assistants who have sufficiently long experience in data annotation for NLP. They were asked to annotate the event labels of each sentence in each document as per the annotation scheme. Subsequently, they were asked to report the main event with more relevant context. While writing context information, they were asked to consider the discourse labels of individual sentences as guiding information.
\subsubsection{Settings:}
 The constants $\alpha$, $\lambda_1$, $\lambda_2$, $k$, and $c$ are optimized using grid search to attain the maximum ROUGE-L + ROUGE-2 score on the development set for both the WCEP and Multi-News datasets collectively.
\subsubsection{Results}
Our results are shown in the Table \ref{tab:rouge_results}. The table show the comparison of our approach with zero/ a few shot results of pre- trained models namely : BART \cite{lewis2019bart}, PEGASUS \cite{zhang2020pegasus}, Longformer-Encoder Decoder(LED) \cite{beltagy2020longformer} and PRIMERA \cite{xiao2021primera}. The comparison is restricted to zero/ a few shot results for a level comparison as our approach is unsupervised as far as summarization is concerned. For experiments we included our approach with main-event bias ($C(S) + \lambda_1 \cdot D(S) + \lambda_2 \cdot B_{\text{main}}(S)$) and without main-event bias ($C(S) + \lambda_1 \cdot D(S)$). Our approach consistently performed better than base models in all the datasets in terms of all the three metrics considered. Base models lack an explicit control of summary content selection and are more probable to hallucinate. On the other hand, our method employs an explicit content extraction mechanism so that re- writing is more guided. This could result in more reliable content and less prone to hallucination. The result increased result in our dataset is obvious as our method is specifically designed the main event with necessary context.  
\begin{figure}[!t]
  \centering
  \includegraphics[height=5.4cm, width=9cm]{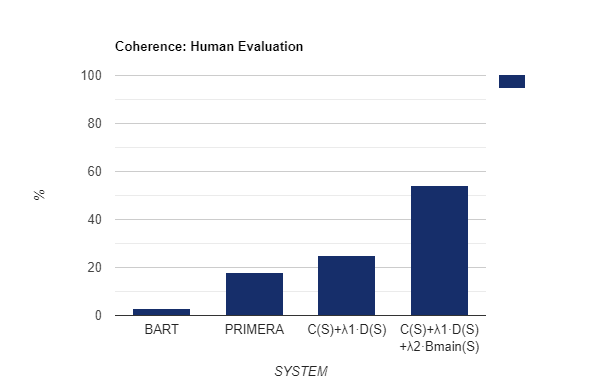} 
  \caption{Human Evaluation for Coherence}
  \label{fig:coh}
\end{figure}

\begin{table*}[ht]
\begin{tabularx}{\textwidth}{X}
\hline\\
\hline\\
\\System Summary \\
\hline
Two women have been charged with criminal damage after climate change protesters threw soup over Vincent van Gogh's painting Sunflowers at London's National Gallery, British police said on Saturday (Oct 15). Van Gogh's Sunflowers, valued at roughly US \$80 million, had no connection to climate change, according to authorities. A video posted by the Just Stop Oil campaign group, which has been conducting protests for the past two weeks in the British capital, showed two of its activists on Friday throwing tins of Heinz tomato soup onto the painting.  \\
\hline
\\Reference Summary \\
\hline
Two women have been charged with criminal damage after climate change protesters threw soup over Vincent van Gogh\u2019s painting Sunflowers at London's National Gallery, British police said on Saturday (Oct 15). A video posted by the Just Stop Oil campaign group, which has been holding protests for the last two weeks in London,\u00a0showed two of its activists on Friday throwing tins of Heinz tomato soup over the painting, one of the treasures of the museum's collection. Police said the two women, aged 21 and 20, would appear later at Westminster Magistrates' Court charged with \"criminal damage to the frame of van Gogh's Sunflowers painting \\
\hline
\hline
\end{tabularx}
\caption{Summaries generated by our system and corresponding reference summaries}
\label{tab:summaries}
\end{table*}
\subsection{Human Evaluation for Coherence}
To evaluate coherence, we selected four human evaluators who are postgraduate students in linguistics. We randomly assembled a sample set of output summaries, comprising 45 summary sets. Each summary set contains summaries generated by the following settings: $BART$, $PRIMERA$, $C(S) + \lambda_1 \cdot D(S)$, and $C(S) + \lambda_1 \cdot D(S) + \lambda_2 \cdot B_{\text{main}}(S)$. The summaries were presented to the evaluators in a randomized order to prevent any potential bias. The evaluators were then asked to identify the most coherent summary among those listed. They were instructed to assess coherence based on discourse connections using linguistic cues and to consider topical continuity between neighboring sentences in the summaries. The results are shown in the Figure \ref{fig:coh}. The settings utilizing large language models (LLMs) were overwhelmingly preferred by the evaluators. This preference is unsurprising, given that LLMs are more reliable for text generation compared to text generation networks with a relatively smaller number of parameters. Even after incorporating LLMs into the text generation process, $C(S) + \lambda_1 \cdot D(S)$, the setting without main event bias was selected a lesser number of times by the evaluators in comparison with $C(S) + \lambda_1 \cdot D(S) + \lambda_2 \cdot B_{\text{main}}(S)$.   Bias term $B_{\text{main}}(S)$  encourages sentences in the extractive summaries to be more semantically aligned with the initially added main event sentence. When the semantic relatedness between extracted sentences is high, a coherent summary can be achieved when re-written by an LLM as semantic relatedness contributes to coherence \cite{aletras2013evaluating}. Table \ref{tab:summaries} contains a sample system-generated summary with high ROUGE scores, along with its corresponding reference summary. The system-generated summary demonstrates a substantial level of content overlap with the reference summary and maintains topical continuity that is quite similar to the reference summary.
\section{Conclusion}
In this current work, we introduced a main event-focused method for multi-document news summarization. Our approach consistently outperformed peer systems in terms of ROUGE metrics. Summaries generated by the method exhibited better coherence, as per human evaluation. Future work could explore improved counterparts for each component of the method, namely main event extraction, context extraction, and rewriting.

\bibliography{aaai24}

\end{document}